\documentclass[sigconf]{acmart}

\usepackage{booktabs} 
\usepackage{subcaption}
\usepackage{csquotes}
\setcopyright{rightsretained}

\copyrightyear{2017} 
\acmYear{2017} 
\setcopyright{acmlicensed}
\acmConference{FDG'17}{August 14-17, 2017}{Hyannis, MA, USA}\acmPrice{15.00}\acmDOI{10.1145/3102071.3102104}
\acmISBN{978-1-4503-5319-9/17/08}

\begin{document}
\title{Mechanics Automatically Recognized via Interactive Observation: Jumping}

\author{Adam Summerville }
\affiliation{%
  \institution{University of California, Santa Cruz }
  \streetaddress{1156 High St.}
  \city{ Santa Cruz} 
  \state{CA} 
  \postcode{95064 }
}
\email{asummerv@ucsc.edu }

\author{Joseph C.\ Osborn }
\affiliation{%
  \institution{University of California, Santa Cruz }
  \streetaddress{1156 High St.}
  \city{ Santa Cruz} 
  \state{CA} 
  \postcode{95064 }
}
\email{jcosborn@ucsc.edu  }

\author{Christoffer Holmg{\aa}rd}
\affiliation{%
  \institution{New York University}
  \streetaddress{6 MetroTech Center}
  \city{Brooklyn} 
  \state{NY} 
  \postcode{11201}
}
\email{christoffer@holmgard.org}

\author{Daniel W.\ Zhang}
\affiliation{%
  \institution{New York University}
  \streetaddress{6 MetroTech Center}
  \city{New York} 
  \state{NY} 
  \postcode{11201}
}
\email{daniel.zhang@nyu.edu}

\begin{abstract}

Jumping has been an important mechanic since its introduction in \emph{Donkey Kong}.  It has taken a variety of forms and shown up in numerous games, with each jump having a different feel.  In this paper, we use a modified Nintendo Entertainment System (NES) emulator to semi-automatically run experiments on a large subset ($\sim$$30\%$) of NES platform games.  We use these experiments to build models of jumps from different developers, series, and games across the history of the console.  We then examine these models to gain insights into different forms of jumping and their associated feel.

\end{abstract}

%
%




\maketitle

\section{Introduction}
Game feel has always implicitly been a part of digital game design ever since Tennis for Two was developed using an oscilloscope at Brookhaven National Laboratory in 1958 \cite{donovan2010replay}. However, it was only recently that Swink \cite{swink2009game} formalized the notion of game feel, defining it as $``$Real-time control of virtual objects in a simulated space, with interactions emphasised by polish$"$ (p. 6).
Spurred on by a increasing interest in formal, quantitative, computational game studies, we conducted a number of investigations into the relationship between the low-level parameters of a game and the phenomenal game feel.

For platform games, jumping is a, if not the, central game mechanic that is highly important for determining the game's feel.
So important is jumping that 2D platform games have been the subject of several studies on this area of game feel and game mechanics.
\citeauthor{aldrich2012guide}, in a number of studies \cite{aldrich2012guide,aldrich2012luigi,aldrich2012super,aldrich2012complete}, investigates how the mechanics and feel of jumping in the Mario series have developed over time.
Relatedly, \citeauthor{fasterholdt2016you} \cite{fasterholdt2016you} show how the feel of jumping in a range of platform games can be described by 21 features, categorized into input, ground movement, jump, air control, jump release, and details.

While these studies are informative for these particular games, their methods typically require either programming tailored to the individual game(s) under study, or painstaking manual analysis of gameplay records, e.g. game-play derived logs and/or video data, mapping controller inputs to events on-screen.

In this paper, we present an automated solution for characterizing jumps and potentially other mechanics in games for the Nintendo Entertainment System (NES).
We scrape data from the video memory of the NES software emulator FCEUX \cite{fceux} and parse the data in terms of the elements on screen, identifying which elements consistently respond to player input. We track these elements and build a jump model by running repeated experiments, capturing the full expressive range of the jump. Using this information, we proceed to investigate commonalities and trends in jumping across games, developers, and game franchises.   We do this in two ways. First, by performing a quantitative analysis using different dimensionality reduction techniques (t-SNE, PCA), we find which jumps are similar to each other and which parameters are truly necessary.  Second, we perform a qualitative analysis of jump arcs across different developers and series.




\section{Related Work}
Swink \cite{swink2009game} performed a qualitative analysis of the jump in \textit{Super Mario Bros.} and speculated about how certain changes to parameters might affect the game feel, but this analysis did not supply any specific parameters.  \citeauthor{fasterholdt2016you} \cite{fasterholdt2016you} show that insights into the link between the low-level implementation of game mechanics and the derived game feel can be used to understand a game's design.
Additionally, they show how formal, comparative studies of several games can be useful to understand the expressive range surrounding particular mechanics.
For instance, this approach can help us understand the limits of when a jump in a platform game is classified as a proper jump in the game's context, and how different instantiations of the abstract jump mechanic can support different design decisions.

\citeauthor{fasterholdt2016you} also tackled the design of jumping across a set of four 2D platformer video games. Their work resulted in a prototyping tool to research game feel specifically for jumps via various tunable parameters.
In contrast, our work spans a larger set of games on one specific console, which allows us to delve in-depth into the exact attributes of game feel for an NES style 2D platformer.

\citeauthor{isaksen2015exploring} \cite{isaksen2015exploring} explored the question of what constitutes difficulty, using parameterized game design to investigate different instantiations of the same set of game mechanics, creating different game variants.
They then investigated the difficulty of these variants using simple simulated players that mimicked simple human player characteristics via variance in input and error rate. 

\citeauthor{ho2016finding} \cite{ho2016finding} recently used search engines and keywords in free-form online textual material describing roguelike games to identify design influences within the roguelike genre.
They automated this analysis process and were able to show how different titles influenced subsequent ones.
Our method is inspired by this approach, but characterizes games not from textual descriptions but rather direct machine observations of the games themselves.




In the following section, we describe the method we use to observe and model jumping mechanics across a range of 2D platform games.

\section{Method}
In this paper, we develop a novel method for automatically identifying and analyzing jumps in 2D platform games, specifically on the NES, as represented by the FCEUX emulator.
We then proceed to automatically study jump mechanics across a range of games developed for the NES and characterize them in terms of the parameters that describe the jump.
As such, this is a two step process which consists of first learning about jumping for each individual game and subsequently comparing the findings across these games.
We first detail our method for automatically identifying and characterizing jumps in individual games using machine learning. Then, we move on to describe how we compared parameters across the selected titles.

It is important to note that we only deal with games for which we know that jumping occurs in the game.
Additionally, we only focus on vertical jumping from a standing position, ignoring any impact of horizontal movement on the jumps and treating only a very specific mechanic and a very particular kind of game feel: vertical jump feel in 2D platform games.
In the following section, we describe how we use insights from design work on jump feel in conjunction with work on \emph{hybrid automata} to automatically identify and learn the parameters of jumps in particular 2D platform games.


\subsection{Jump Automaton}
\citeauthor{swink2009game} describes how the movement of characters controlled by the player in 2D platform games~\cite{swink2009game}, such as \emph{Mario} \cite{supermariobros1985miyamoto}, can be characterized using two main representations: finite state machines and \emph{attack, decay, sustain, and release} envelopes.
Envelopes are a specialized notation for how individual continuous variables change according to simple (fixed) state machines, borrowed from the domain of music synthesis.
Swink used them to great effect, illustrating their explanatory power even when drawn freehand from naked observation with no access to the code or memory of the game in question.  

Swink used state machines to describe discrete behaviors and envelopes for continuous ones; as it turns out, the discipline of control theory has a useful model which captures both aspects of these mixed discrete and continuous systems: \emph{hybrid automata}.
Hybrid automata are finite state machines augmented with continuous variables, where those variables update continuously at different rates in different states.
Transitions between states can occur when those variables cross thresholds (or exogenous inputs or events arrive) and the act of transitioning between states may cause instantaneous changes to variable values~\cite{alur1993hybrid}.

\begin{figure} 
  \centering
  \includegraphics[width=\columnwidth]{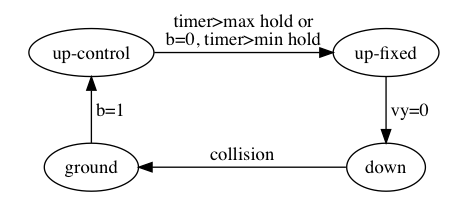}
  \begin{tabular}{l l}
    Parameters & Fasterholdt \emph{et al.}\\
    \hline
    \emph{max hold} & Button hold time; determines \emph{max jump time}\\
    \emph{min hold} & Button hold time; determines \emph{min jump time}\\
    \emph{up-control gravity} & Inverse of \emph{additive jump force} \\
    \emph{up-control multiplier} & Not present \\
    \emph{up-control reset} & \emph{Takeoff velocity vertical} \\
    \emph{up-fixed gravity} & Inverse of \emph{additive jump force} \\
    \emph{up-fixed multiplier} &  Not present \\
    \emph{up-fixed reset} & Determines \emph{instant brake} \\
    \emph{down gravity} & \emph{Gravity} \\ 
    \emph{down multiplier} & Not present \\
    \emph{down reset} & Not present; generally 0 
  \end{tabular}
  \caption{Our jumping automaton and corresponding Fasterholdt \emph{et al.}\ parameters.}
  \label{fig:ha}
\end{figure}

The representational power of hybrid automata---as well as prior work in learning parameters on fixed hybrid automata~\cite{yordanov_parameter_2008}---led us to (manually) define an automaton suitable for modeling Mario's jump.
The choice of Mario was motivated by the assumption that, on the NES platform,  Mario's jumps are as complex as any other, ignoring double or triple jumping.
Given this assumption, we believed that any state machine sufficient to describe Mario's jump would be sufficient to describe jumping in most NES 2D platform games.
Looking closely at Mario's jump (and guided by the discussion in~\cite{swink2009game}), we composed a state machine out of four discrete states: \emph{ground}, \emph{rising with jump-button}, \emph{rising without jump-button}, and \emph{falling} (see Fig.~\ref{fig:ha}).
Mario transitions from \emph{ground} to \emph{rising with jump-button} when the jump button is pressed (simultaneously receiving a large upward velocity); from \emph{rising with jump-button} to \emph{rising without jump-button} when the button is released and the player yields control over the jump; from either \emph{rising} state to \emph{falling} when reaching the apex of the jump; and from \emph{falling} to \emph{ground} when touching solid ground again.

Each state has a number of different parameters that determine the continuous evolution of the vertical position \(y\) and velocity \(v\); in the end, Mario's vertical position when in a given state \(s\) is defined as the following function of time and the values of \(y\) and \(v\) as of entering the state:
$$y_t = y_0 + (v_{y,s} + m_{0,s}v_0)t + a_{y,s}t^2$$
where $y_t$ is the position at time $t$, $v_{y,s}$ is the current reset velocity, $m_{0,s}$ is a multiplier for the initial velocity $v_0$ upon entering this state, and $a_{y,s}$ is the current value of gravity.
This captures cases where the velocity is unchanged on entering the state; when it is reset to a constant value on entering the state; or when it is set to a multiple of the original velocity on entering the state.
We also assume every state has a constant acceleration (possibly zero).

Besides the three per-state parameters described above, we learn two global parameters governing the transitions from \emph{rising with jump-button} to \emph{rising without jump-button}: the \emph{min hold} and \emph{max hold} duration.
Any button press of duration less than the \textit{min hold} will produce the exact same jump as the button held for the \textit{min hold}.
To illustrate, in \emph{Metroid}, all jumps with a button press between 1-10 frames are identical, but a jump of 11 frames is different.
Similarly, any button press longer than the \textit{max hold} will be equivalent to holding the button for exactly the \textit{max hold}.

Unsurprisingly, we found that simpler jump mechanics such as fixed-animation jumps or jumps where the ascending and descending gravity are the same are specializations of this same state machine.
For example, a fixed-animation jump like that of \emph{Castlevania's} Simon Belmont has equal values for the \emph{min} and \emph{max hold}; and the main distinction between the jumps of Mario and \emph{Metroid's} protagonist Samus Aran is that Mario's \emph{rising without jump-button} and \emph{falling} modes are distinct, while Samus's have identical parameters.
We proceed under the assumption that this state machine is sufficient to describe jumps in the majority of 2D platform games on the NES.  

The parameters described above were chosen in part to overlap with those identified in \cite{fasterholdt2016you}.
While we started with the full set of parameters, we decided to narrow our scope to a subset that were relevant to standing vertical jumps.
We made this decision to minimize the number of variables implicated, to reduce the complexity of the state machines, and to simplify our experimental procedure for this initial study.
In future work, this could be expanded to cover other kinds of jumps or other game mechanics like flying.
In most cases elaborating on the automaton could suffice on its own without substantial algorithmic changes: for example, \emph{Kirby}-style infinite jumping could be obtained by a loop from \emph{rising without jump-button} and \emph{falling} back to \emph{rising with jump-button}; and the flight of Raccoon Mario could be modeled by adding \emph{floating downwards} and \emph{flying upwards} states.

\subsection{Experimental Methodology}
The automaton described in fig.~\ref{fig:ha} represents our general model of jumping that we assume can cover jumping in NES games adequately.
In this model, we define the following experimental methodology to learn an instantiation for each 2D platform game included in our dataset:
\begin{enumerate}
\item Data acquisition: Acquire raw sprite position information
\item Sprite tracking: Track the sprites and determine which is the player
\item Jump mode separation: Separate the track into different modes of jumping such that we can model each jump type individually.
\item Parameter Learning: Learn the parameters of the modes to get the most accurate representation of each jump mode in the general automaton framework.
\end{enumerate}
These steps are described in detail below.

\subsubsection{Data Acquisition}
Tom7's \textit{The glEnd() of Zelda} ~\cite{murphy2016glend} is our inspiration for the data acquisition step. 
Using an open source emulator, he was able to extract data directly from the Picture Processing Unit (PPU) of the Nintendo Entertainment System (NES).
He used memory manipulation to position the character in different places and see if there was gravity, which tiles were solid, etc.
We have a more precise goal of determining the exact structure of a given jump, not broadly whether jumping (or gravity in his work) exists.
His work also requires per-game knowledge about the memory structure of the game for manipulation, whereas this work makes no assumptions about the memory structure. 
By directly accessing the sprite table of the PPU, we are able to extract perfectly precise information about the positions at each frame, without relying on any computer vision techniques. 

To acquire the data, a human must play the game and record their actions to get the game into a safe state from which we can observe a jump. To clarify, a safe state is a state where the player character will not die within the span of several seconds.
For most games, this is as simple as pressing start, but some other games start with the player situated in a place with dangerous enemies and/or environmental hazards, which will interrupt the experiments,  e.g. \emph{Super Mario Bros.} starts in a safe area while \emph{Metroid} starts such that the player must move a few tiles to the left or right to enter a safe state. From this safe state, we run a number of trials via the following steps:

\begin{enumerate}
\item Hold down the jump button for $k$ frames
\item Wait for $j$ frames
\item Reset to the safe state, incrementing to$k$ to $k+1$ frames, and go back to (1).  After $n$ experiments, exit.
\end{enumerate}

During the experiments we record all sprite positions per frame.  For this work, $k =1, j = 120,$ and $n=120$.  We note that we only analyze jumps in the absence of other mechanics.  This in part comes from Swink, who only considered 1 dimensional envelopes (e.g. horizontal or vertical speed).  While this is not how players actually encounter the mechanics (i.e. the player does not only ever move in one dimension at a time), for most games, the mechanics that govern one direction are orthogonal to the other (i.e. a standing jump is identical to a running jump).  For the games where this is not the case (say \textit{Super Mario Bros.}) it remains as future work to fully extract all of the jump mechanics.

To track the sprites we extract all of the sprite information from the sprite table of the PPU.
The sprite table contains up to 64 entries, with each entry consisting of the index of the bitmap to look up in a table, the $x$ and $y$ coordinates of the sprite, whether the sprite is horizontally and/or vertically flipped, whether the sprite is in the foreground or background, and the color palette of the sprite.
The combination of sprite id, color palette, foreground, and flipping information define a unique sprite, so each set of those is treated as a unique sprite.  

It is important to note that each of these sprites are actually only $8\times8$ or $8\times16$ pixels and most characters are made up of multiples (e.g. Super Mario is made up of $8$ $8\times8$ sprites).
To account for this, we perform a filtering step.  We look at the entirety of the experiment and keep track of which sprites are touching at each time step.  For each pair of sprites this will give us the probability of the two sprites touching, $\mathrm{p}(x,y)$, as well as the probability of a given sprite being on the screen, $\mathrm{p}(x)$.  From these we calculate the Normalized Pointwise Mutual Information (NPMI) \cite{summerville2017what}, a measure of how likely two events are to correspond.
A value of -1 means the events are perfectly anti-correlated, 0 that they are independent, and 1 that they always co-occur.  Any two sprites that pass a threshold of 0.1 are chosen to be merged.
At each time step we find all pairs that should be merged and merge them into disjoint sets, with each set representing a fully merged sprite composed of many sub-sprites.

\subsubsection{Sprite Tracking} 
Given the merged sprites (hereafter referred to just as sprites), we need to track them across multiple time steps.  At the beginning of the experiment, there are no tracks, so each sprite on the screen initiates a new track.  In subsequent timesteps, we need to determine which track each sprite belongs to, if any at all.  Standard target tracking algorithms assume a maneuver model that has inertia \cite{kalman,imm}, but game characters can exhibit non-physical dynamics, so such algorithms are unsuitable.  To allow for instantaneous changes in a sprite's movement direction, we make no assumptions about the underlying movement model.  For each pair of sprite and track ($<$sprite, track$>$), we calculate the Euclidean distance between them.  We assume that a tracked sprite is equally likely to move in any direction, but is most likely to be close to its last known position.  We determine the likelihood of a sprite belonging to a given track given a Normal distribution, $\mathrm{N}(0,8)$, based on the distance, $d$, giving us a likelihood, $L = \mathrm{N}(0,8)(d)$.  We chose 8 pixels as the standard deviation due to the fact that it is the standard width of the sub-sprites.  Given the likelihoods for each $<$sprite,track$>$ pair, we then construct a bi-partite graph, with each sprite on one side and each track on the other, as well as a track initiation node for each sprite.  The edges between each pair is set to the previously calculated likelihood, and each sprite is connected to its track initiation node with an edge weight corresponding to 5 sub-sprite widths, i.e. 40 pixels.  40 pixels was chosen as we do not reasonably believe that a difference that large represents a mechanic other than standard motion, i.e. teleportation or creation of a new sprite.  We then perform a max-weight matching to find the optimal assignment of sprites to tracks.  In many games, a sprite might flicker to show for some mechanical purpose (e.g. to indicate invicibility), and to account for this, we allow a track to coast for 4 timesteps with no updates; if a track has had no new data points after 4 timesteps, it is removed from the current set of active tracks.  Once the complete tracks are formed, they are then used as the input for the jump finding code.

\subsubsection{Jump Mode Splitting}
Given the tracked data, we must filter out all sprites other than the player's character.  To do this, we do a simple jump mode fitting, as follows:

\begin{enumerate}
\item If a sprite is the player character, it starts on the ground, $y_g$
\item It will go up at some point, so there must exist an apex, $y_a$, $y_a > y_g$
\item After rising, it must go down, so all points after the apex should be lower
\item It finishes on the ground, plus or minus a pixel (the reasons for which are addressed below)
\end{enumerate}

Any sprite track that does not have these characteristics is thrown away.  Given the filtered sprites, we must then determine the relevant modes found in the jump.  First, we find whether there is any \textit{up-control} mode.  Throughout the experiments we keep increasing the length of the button press, but this does not necessarily have any effect on the duration of the jump.  A jump that has equal \textit{min} and \textit{max} hold has no \textit{up-control} mode, and is a \textit{fixed} jump, as opposed to a \textit{controlled} jump.  Given these two classes of jumps, we need to determine when the jump transition between modes from \textit{ground} $\Rightarrow$  \textit{up-control} and/or \textit{up-fixed} $\Rightarrow$ \textit{down} $\Rightarrow$ \textit{ground}.  We define those transitions as:

\begin{itemize}

\item \textit{ground} $\Rightarrow$  \textit{up-control} or \textit{up-fixed} : It is within the first $b$ frames of the experiment and $y_{t} > y_{t-1}$

\item   \textit{up-control} $\Rightarrow$  \textit{up-fixed} : Either both, $t_s >  $ \textit{min hold}  and the jump button is released, or $t_s > $ \textit{max hold}, where $t_s$ is the time in the state  

\item \textit{up-fixed} $\Rightarrow$ \textit{down}  : If $y_{t-1} \leq y_t < y_{t+1}$, i.e. $y_t$ is the apex of the jump.

\item \textit{down} $\Rightarrow$ \textit{ground} : If $y_{t-1} > y_t$  and $y_t = y_0 \pm 1$  - The $\pm 1$ is due to the fact that player characters in some games actually land 1 pixel above or below the ground and have a short animation that transitions them back onto the ground.

\end{itemize}

\begin{figure}
\centering
    \begin{subfigure}[t]{0.46\textwidth}
        \centering
        \includegraphics[width=1\textwidth]{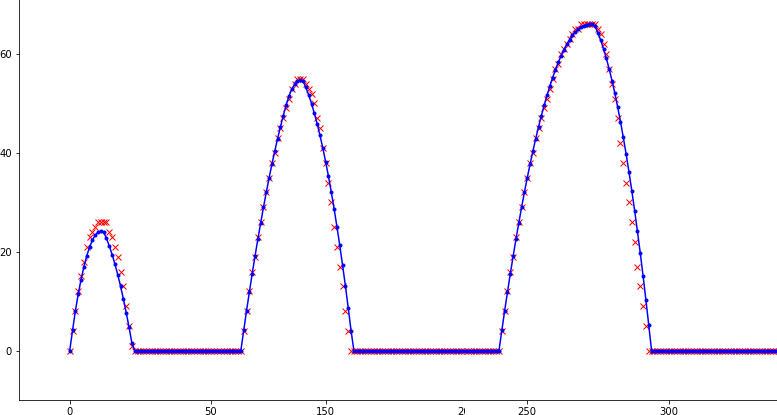}
        \caption{\emph{Super Mario Bros.}}
    \end{subfigure}%
     
    \begin{subfigure}[t]{0.46\textwidth}
        \centering
        \includegraphics[width=1\textwidth]{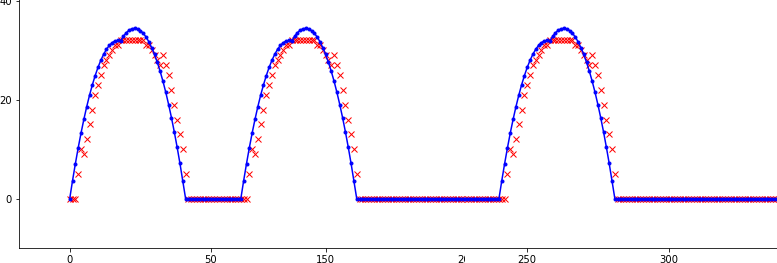}
        \caption{\emph{Castlevania}}
    \end{subfigure}
    
    \begin{subfigure}[t]{0.46\textwidth}
        \centering
        \includegraphics[width=1\textwidth]{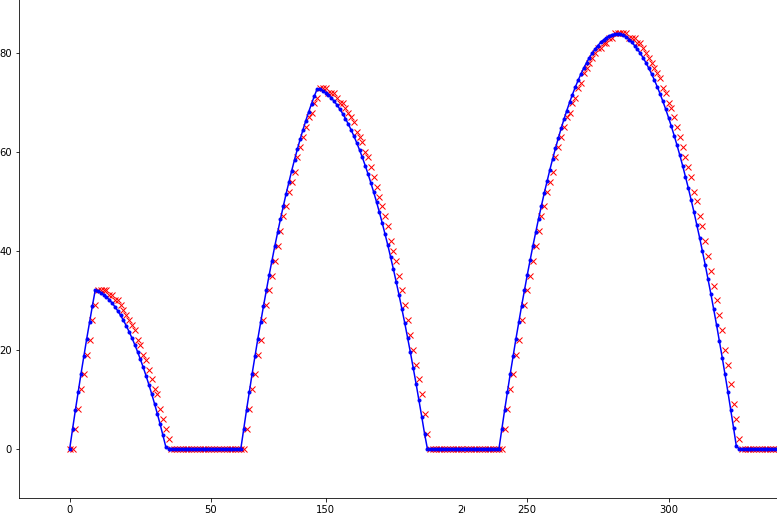}
        \caption{\emph{Metroid}}
    \end{subfigure}
\caption{Three jumps representing the minimum button hold duration, median button hold duration, and maximum button hold duration for \emph{Super Mario Bros.},\emph{Castlevania}, and \emph{Metroid}. The true values are red X's and the predicted values are the blue .'s and lines.   } The horizontal axis is time, not horizontal position.
\label{fig:3jumps}
\end{figure}

\begin{figure}[h!]
\centering
        \includegraphics[width=0.46\textwidth]{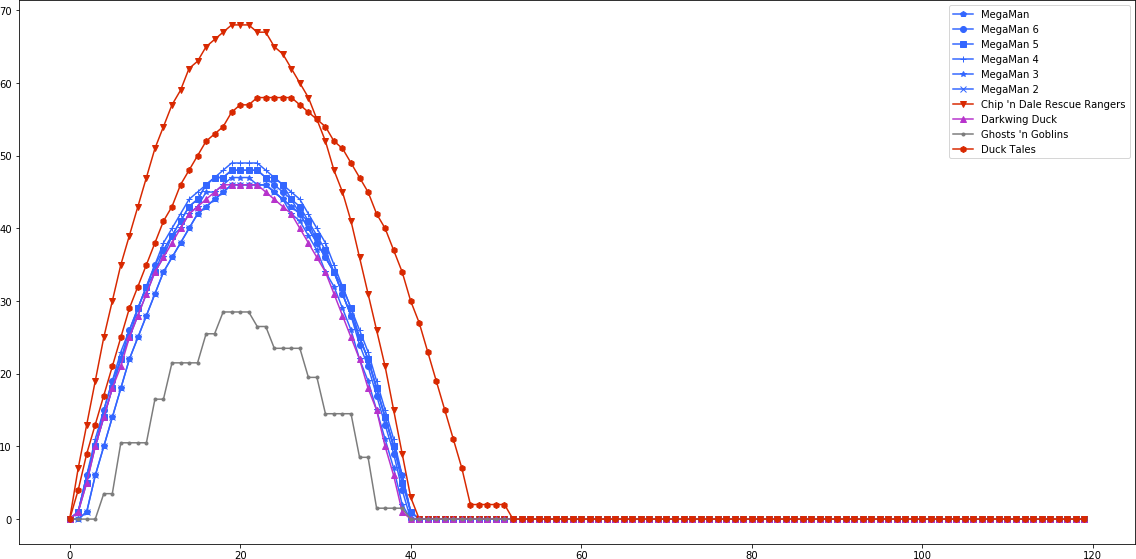}
        \caption{A comparison of jumps across games developed by Capcom on the NES.  }
        \label{fig:CapcomComp}

\end{figure}

Note that while our position data were integral (pixels), many games internally represent characters' positions with some sub-pixel precision; we ignore these cases, which introduces some amount of non-random error into our recorded positions, velocities, and accelerations.  Furthermore, some games do not have any sort of physics model and instead have very discrete fixed paths that they follow (see the \emph{Ghosts 'n Goblins} jump in figure \ref{fig:CapcomComp}).  Furthermore, the nature of our tracking is based on the bounding boxes, which occasionally have drastic changes during the course of a jump, (e.g. Link in \emph{The Legend of Zelda II} tucks his legs, which produces a sharp jolt in the jump, seen in figure \ref{fig:NESComp}).  

To handle the fact that we might have non-random errors in our dataset, we used a linear Support Vector Regression (SVR) \cite{svr}, a variant of linear regression that stipulates that all of the points used for training must fall within an $\epsilon$ band around the regression, with a penalty applied to each point that falls outside of this band all while minimizing the size of the learned weights.  This has the effect of ignoring outliers from the aforementioned error sources.  To find the parameters of the jump we solve the parameters for the equation:

$y_t = y_0 + (v_y + m_0v_0)t + a_yt^2 $

Where $y_t$ is the position at time $t$, $v_y$ is the initial velocity, $m_0$ is a multiplier for the initial velocity upon entering the state, $v_0$, and $a_y$ is the acceleration.  $y_t$, $y_0$, $v_0$, and $t$ are observed variables, and we learn $y_0$, $v_y$, $m_0$, and $a_y$.

\section{Pragmatic Concerns}

We note that while most jumps are ably handled by this system, a number of issues can arise in practice.
The largest source of difficulty are the jumps that are not governed by physics.  Some, such as \emph{Castlevania}, follow a preset trajectory that is non-physical, i.e. it is only parabola-like, while others follow a parabola but have a ``stair-step'' pattern (see \emph{TMNT} in figure \ref{fig:KonamiComp}).
Though the \emph{Castlevania} style leads to a jump model that is close, with systemic errors (see figure \ref{fig:3jumps}), the stair-step model represents a more severe issue.
In general, we look for a segment of 3 frames in which the middle frame is greater than or equal to its neighbors, but in these models, the character might be at the same $y$ position for 5 or 6 frames.
To account for this, the \emph{down} state is only entered if the sprite has begun to fall, i.e. the previous frame can be equal to the apex, but the next frame must be lower; however, this is not infallible as jitter in the position due to animations can result in momentary decreases.
While this issue does not arise in any of the games used for this analysis, it is a potential source of future error.  

Another concern is that some games have oddities upon landing, with
some games having characters clipping momentarily into the ground; the game with the most pronounced effect is \emph{Darkwing Duck}, in which the player character clips 8 pixels into the ground while falling, before snapping back to the ground.
We clamp the lowest $y$ position to the initial ground $y$ position for parameter learning, but this results in a model that is an over-approximation.  Other games do not end on the ground such as \emph{Adventure Island II}, which finishes 1 pixel above the ground.
It then enters into an animation loop that stays up 1 pixel for 8 frames, goes down to the ground for 8 frames, and repeats.
To account for this, we added the $\pm 1$ pixel slack, but in general this seems to be a difficult pattern to account for, especially in conjunction with stair-step jumps that might stay at an arbitrary height for an arbitrary number of frames.

\section{Analysis}
We performed our analysis on 48 games from the NES library.
While not an exhaustive examination of platformers on the NES, we have 30\% coverage of platformers released for the system.
Due to the fact that some games have different dynamics based on the selected player character per game, we learned 52 characters.
As such, the analysis presented below is done at the level of individual NES characters, where a few of them happen to appear in the same game.

We aim to answer four main questions:
\begin{itemize}
\item How different are 2D platform games across the NES system, and can jumps be grouped into different categories?
\item Do games in the same franchise or by the same developer or publisher exhibit similar jump characteristics?
\item Does the style of jumping in NES games evolve over the life-time of the platform, as measured by publication date?
\item Does our method capture differences in jump styles that can be recognized and explained qualitatively, i.e. does our method have face validity.
\end{itemize}
In order to address these questions, we first conduct a quantitative exploratory analysis of the collected data and then a qualitative analysis of these results.

\subsection{Characterizing NES 2D platformer jumps}

\begin{figure}[!]
        \centering
	        \includegraphics[width=0.4\textwidth]{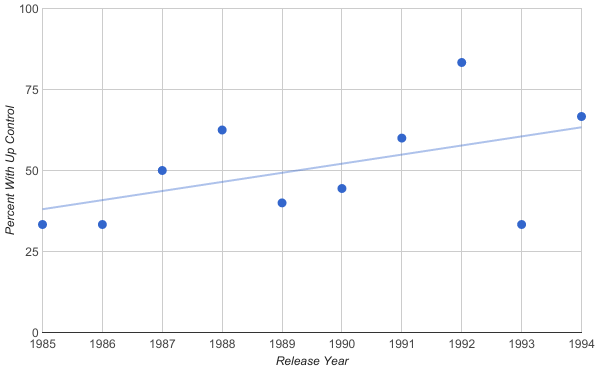}
        \caption{Percentage of games released with up-control per release year.}
        \label{fig:controlVtime}
\end{figure}

Figure \ref{fig:controlVtime} shows the percentage of games released with up-control per release year.  \emph{Super Mario Bros.} brought with it the advent of the up-control jump in 1985.  While the controlled height jump was not universally adopted, we note a clear trend with more games adopting it.

Figure \ref{fig:3jumps} shows the shortest, median, and longest jumps possible for three games. We note that the horizontal axis for all shown jumps is time, not horizontal distance. These games are chosen due to the fact that each represents a different model of jumping. \emph{Super Mario Bros.} has a period of control, a reset after button release, and a subsequent increased gravity for falling.  \emph{Castlevania} is unaffected by the length of the button press and does not follow a physical model, instead having the character follow a preset trajectory with a very long hover.  \emph{Metroid} has an instantaneous transition from jumping to falling upon button release, resulting in the sharp trajectories on the left, with only the longest jump representing a smooth arc.  We can see that our model captures the dynamics of  \emph{Super Mario Bros.} and \emph{Metroid} well, but has difficulty with  \emph{Castlevania} due to the fact that its arc is not a quadratic parabola.

\subsection{Dimensionality Reduction Using Principal Component Analysis}
In this subsection we focus on dimensionality reduction and clustering in order to explore whether the games in our dataset can be divided into categories.
Not all parameters are meaningfully learned for all games, due to the fact that some games do not have any jump control. Accounting for this, we used the shared subset of the learned parameters:
\emph{min duration, max duration, down reset, down multiplier, down gravity, up-fixed reset, up-fixed multiplier, up-fixed gravity} as well as parameters for the initial reset and gravity (from \emph{up-control} if the jump has control, from \emph{up-fixed} if it does not) and an indicator variable for whether the jump has control or not.

\begin{figure}[!]
        \centering
          \includegraphics[width=0.48\textwidth]{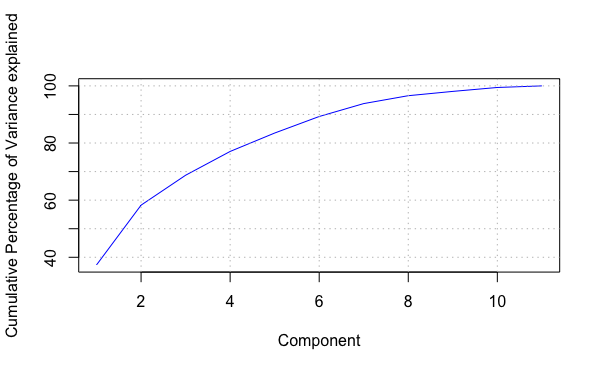}
        \caption{Components found from Principal Component Analysis of the features shared across all game characters. 4 components are enough to account for more than 75\% of the variation.}
        \label{fig:jumpdata_scree}
\end{figure}

First, we perform dimensionality reduction through unrotated principal component analysis (PCA). This allows us to measure and visualize how, if at all, a lower number of latent components can account for the variation in jumps across the games. Figure~\ref{fig:jumpdata_scree} shows a scree plot of the found components, indicating that \~4 components are sufficient to account for most (77.05\%) of the variance in the data set, with subsequent components contributing little individually.  The cumulative contribution of each component can be seen in figure \ref{fig:jumpdata_scree}.    The proportion of how much each parameter contributes to the top 4 components can be seen in table \ref{tab:contribution}.  Generally, we can think of the 4 components as:
\begin{itemize}
\item  The component based on whether the jump has control or not, and the associated up-fixed parameters.
\item The base components, the initial up parameters, and the down gravity
\item The min hold
\item The down multiplier
\end{itemize}

\begin{table}[ht]
\centering
\begin{tabular}{rrrrrrrrrrrr}
  \hline
 & Comp.1 & Comp.2 & Comp.3 & Comp.4 \\ 
  \hline
  max hold & 17.98 & 1.22 & 1.89 & 0.09 \\ 
  min hold & 0.38 & 0.06 & 61.72 & 19.51 \\ 
  initial gravity & 8.20 & 20.28 & 0.79 & 2.10 \\ 
  initial reset & 6.02 & 24.15 & 2.12 & 2.31 \\ 
  up-fixed gravity & 0.17 & 19.09 & 9.60 & 15.88 \\
  up-fixed multiplier & 13.57 & 3.50 & 14.16 & 2.00 \\ 
  up-fixed reset & 19.51 & 0.16 & 0.20 & 0.90 \\ 
  down gravity & 0.82 & 21.78 & 0.33 & 0.04 \\
  down multiplier & 5.48 & 0.25 & 3.52 & 56.56 \\  
  down reset & 13.08 & 4.89 & 0.98 & 0.60 \\ 
  has control & 14.81 & 4.61 & 4.69 & 0.00 \\
  \hline
\end{tabular}
\caption{The contribution of the features shared across all game characters to the four most important latent components.}
\label{tab:contribution}
\end{table}

\subsection{Dimensionality Reduction using t-SNE}
\begin{figure}
\centering
  \hspace{-2cm}\def\svgwidth{0.38\textwidth}
    \input{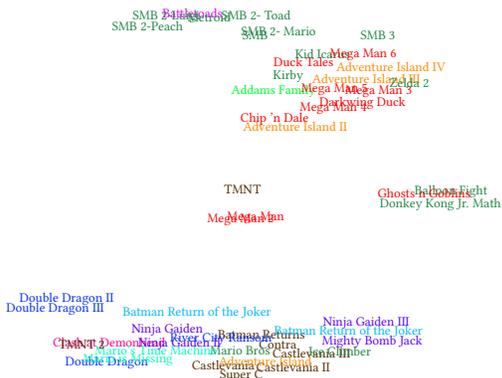}\vspace{-4\baselineskip}
  \caption{t-SNE embedding of the jumps. The color represents the developer, Nintendo: Dark Green, Konami: Brown, Hudson Soft: Orange, Capcom: Red, Rare: Fuchsia, Radical Entertainment: Teal, Technos Japan: Purple, Sunsoft: Light Blue, Tecmo: Dark Blue, Vic Tokai: Pink}
  \label{fig:tsne}
\end{figure}

Figure \ref{fig:tsne} shows a different dimensionality reduction, t-distributed stochastic neighbor embedding (t-SNE).  t-SNE operates by constructing a probability distribution over pairs in the dataset such that similar items have a higher probability and then learns a $k$-dimensional ($k =2$ in this case) mapping such that more similar items are closer to each other.  The upper cluster represents jumps in the \emph{Super Mario Bros.} family, i.e. up-control, while the lower cluster are fixed jumps.  We notice that most of the Capcom games are clustered, with the latter Mega Man and the Disney licensed games grouping together.  The right cluster is comprised of games that have fixed jumps and are very short, both in height and duration.  Konami has a very strong, identifiable house style, with all but one of their platformers being tightly grouped together.  The outlier, \emph{TMNT}, has a very different jump, which can be seen in figure \ref{fig:KonamiComp}.  We notice that some series have large differences that occur part way through the series, such as the difference between \emph{Double Dragon} vs \emph{Double Dragon 2 \& 3},\emph{ Mega Man 1\&2} vs \emph{Mega Man 3-6}, and \emph{Adventure Island} vs \emph{Adventure Island 2-4}.  Adventure Island represents a very large difference, changing the entire modality of the jumping by adopting the Mario style controlled jump, while the Double Dragon series is more of a tweaking of the parameters.  The Mega Man games operate somewhere in between, and this is discussed in more depth below.

\subsection{Clustering}
In order to investigate whether games can be grouped together, and whether there are patterns in terms of publishers and years of release, we cluster games using K-Means.
We calculate the within cluster sum of squares for a number of Ks ranging from 2 to 15 and determine that after 3 clusters no substantial improvement is seen.

The three clusters which emerge can be interpreted as three styles in terms of jump feel:
The red cluster 1 contains jumps with some amount of air control, exemplified by Mario in the Super Mario Bros. series.
The green cluster 2 contains jumps that have a high degree of air control to the extent of making the character jumps feel floaty. The only two Nintendo characters that exhibit this jump are Luigi and Peach from Super Mario Bros. 2. These jumps are graphed in contrast to other Nintendo jumps in Figure~\ref{fig:MarioComp} where it is evident that these characters have much longer, flatter jump curves than the other Nintendo characters from the Mario franchise.
Finally, the blue cluster 3 contains tight fixed jumps with many of the characters stemming from games ported from arcade titles. For the arcade games this jump style may have been motivated partially by technical constraints, but we speculate that it was carried over to pure console games as a design choice or convention.
The categories are further analyzed in Section~\ref{sec:qualitative}.

\begin{figure*}[!]
\centering
        \includegraphics[width=1\textwidth]{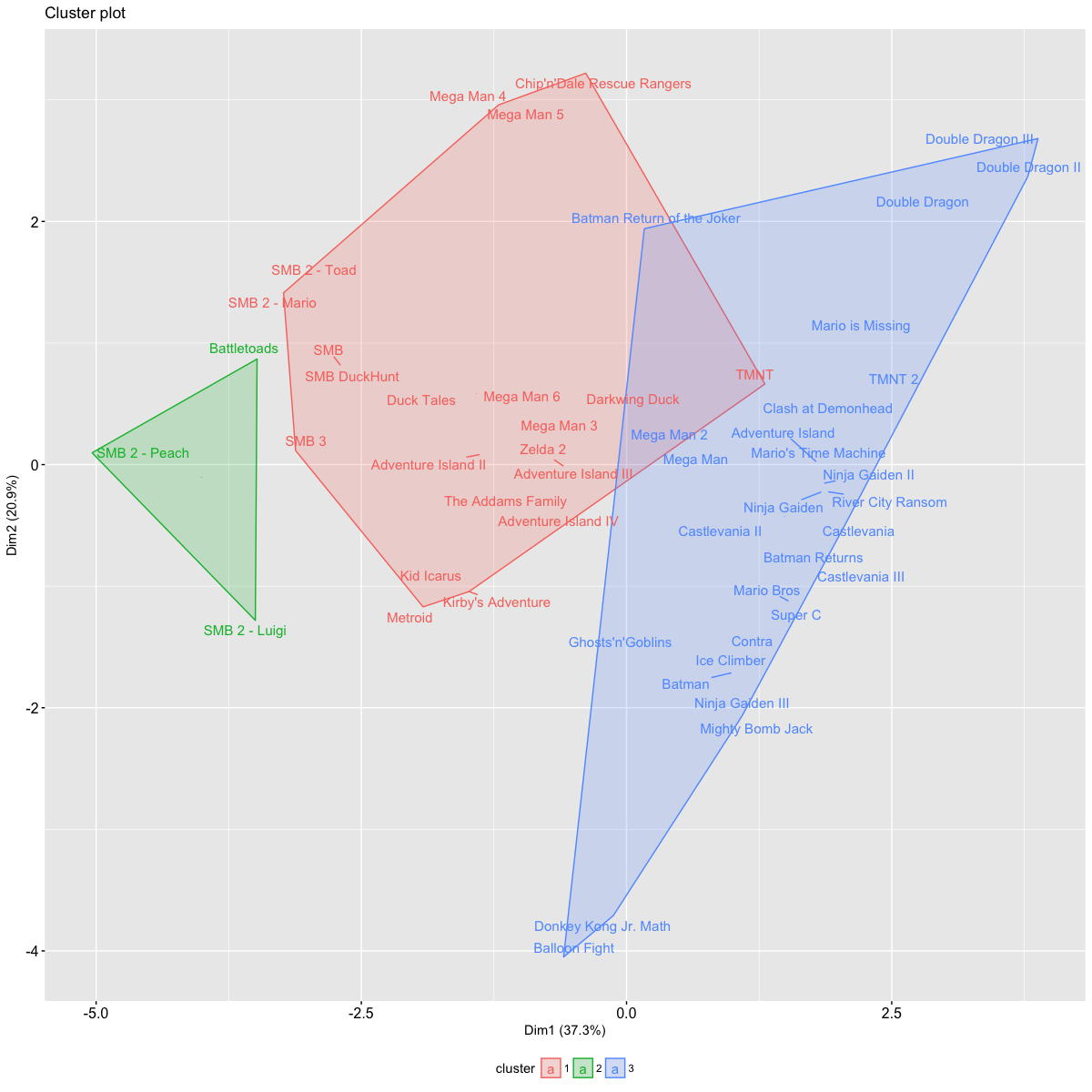}
        \caption{Clusters of games, based on the dataset of features shared by all games.}
        \label{fig:kmeans}
\end{figure*}

When broken down across publishers a number of interesting patterns appear.
Capcom and Nintedo were both prolific in publishing games with jumps from the red cluster in Figure~\ref{fig:kmeans}, i.e. games with a resonable amount of jump control, akin to Super Mario Bros.
As this is the largest and most dispersed cluster it covers a range of different control levels from Mario in Super Mario Bros. 2 to Teenage Mutant Ninja Turtles.
Konami, on the other hand, are chiefly represented in the blue cluster (along with a smaller number of Nintendo games) with titles that feature characters with tight, fixed jumps.
This cluster also contains most of the jumps that were ported from arcade games.
Aside from Battletoads, which is closer to a classic Super Mario Bros. style jump, the two other characters in the third, smallest, green cluster are both from Nintendo's Super Mario Bros. 2 and could be considered experimental characters.

\begin{figure}[h!]
\centering
        \includegraphics[width=0.46\textwidth]{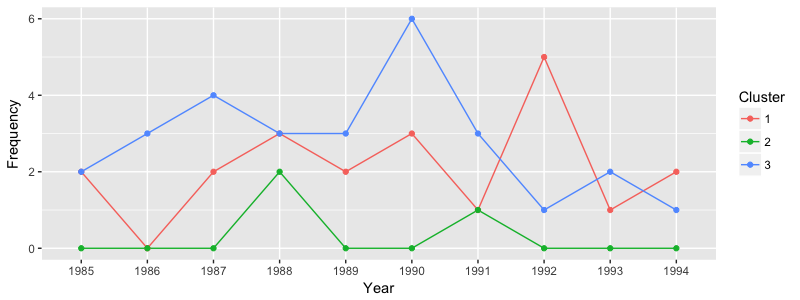}
        \caption{Cluster frequencies by year}
        \label{fig:clusterYear}
\end{figure}

When viewed over time, as displayed in Figure~\ref{fig:clusterYear}, we see that in this set, game characters with tight, fixed jumps are more common in the earlier years of the NES console while higher degrees of air control become more common over time.

Altogether, dimensionality reduction and cluster analysis suggest that 2D platform games on the NES can be grouped into at least three categories, in terms of jump feel:
\begin{enumerate}
\item Medium length, Mario-style jumps with some amount of air control (21 jumps).
\item Experimental, floaty jumps with a large amount of air control (3 jumps).
\item Tight, fixed-length jumps (28 jumps).
\end{enumerate}
There are indications in our dataset that tight jumps are more common in the early life of the NES console while Mario-style jumps become more common later, but this is hard to say conclusively due to the relatively small number of observations and the non-random sampling of the character jumps included in our dataset.

\subsection{Qualitative Analysis}
\label{sec:qualitative}

\begin{figure}[h!]
\centering
        \includegraphics[width=0.46\textwidth]{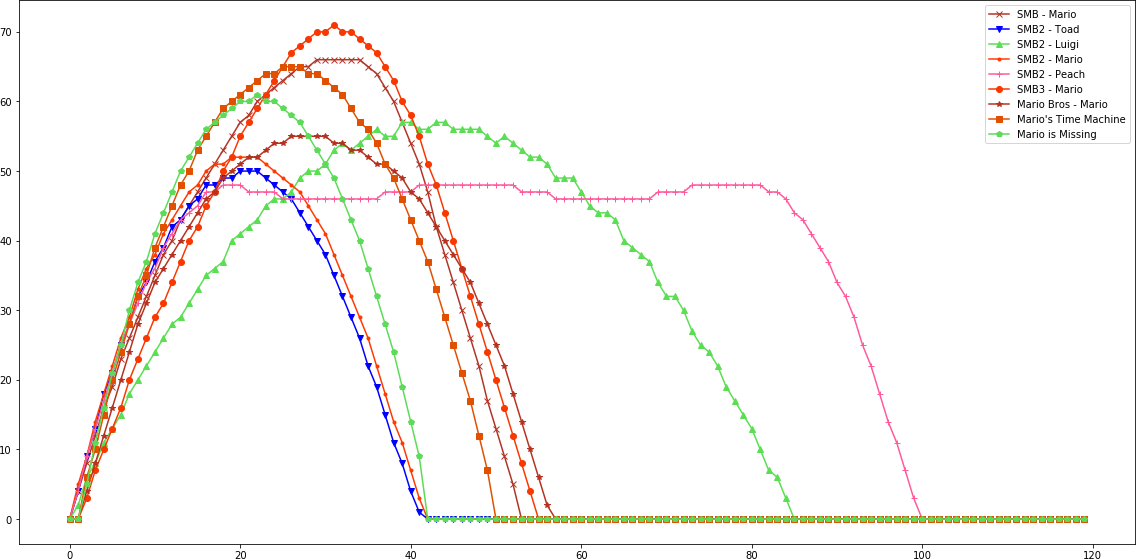}
        \caption{A comparison of jumps across the games in the Mario series on the NES.  }
        \label{fig:MarioComp}
\end{figure}    

\begin{figure}[h!]
\centering
        \includegraphics[width=0.46\textwidth]{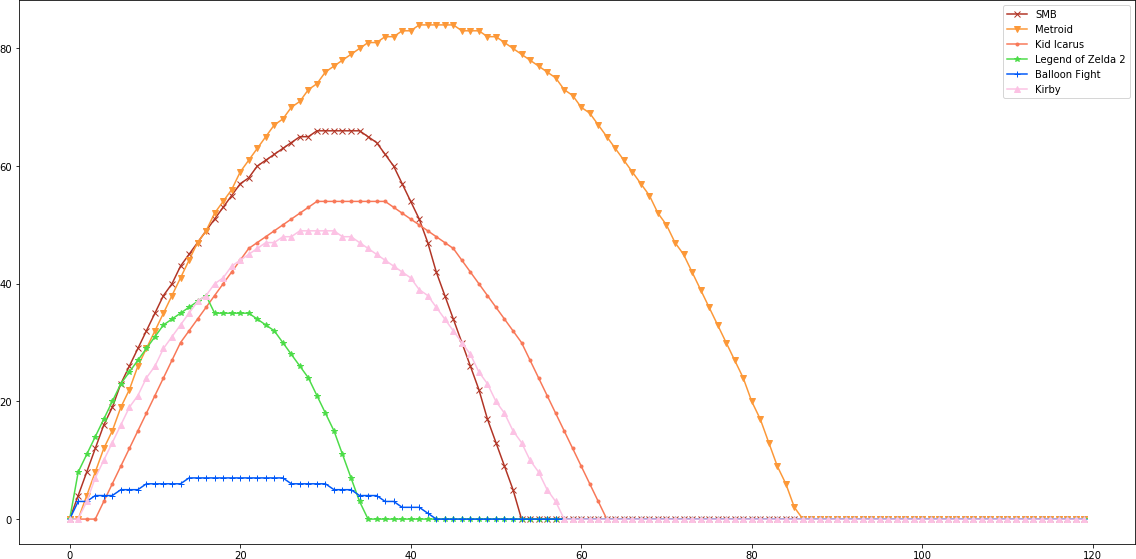}
        \caption{A comparison of jumps across games developed by Nintendo on the NES.  }
        \label{fig:NESComp}

\end{figure}    

\begin{figure}[h!]
        \centering
        \includegraphics[width=0.4\textwidth]{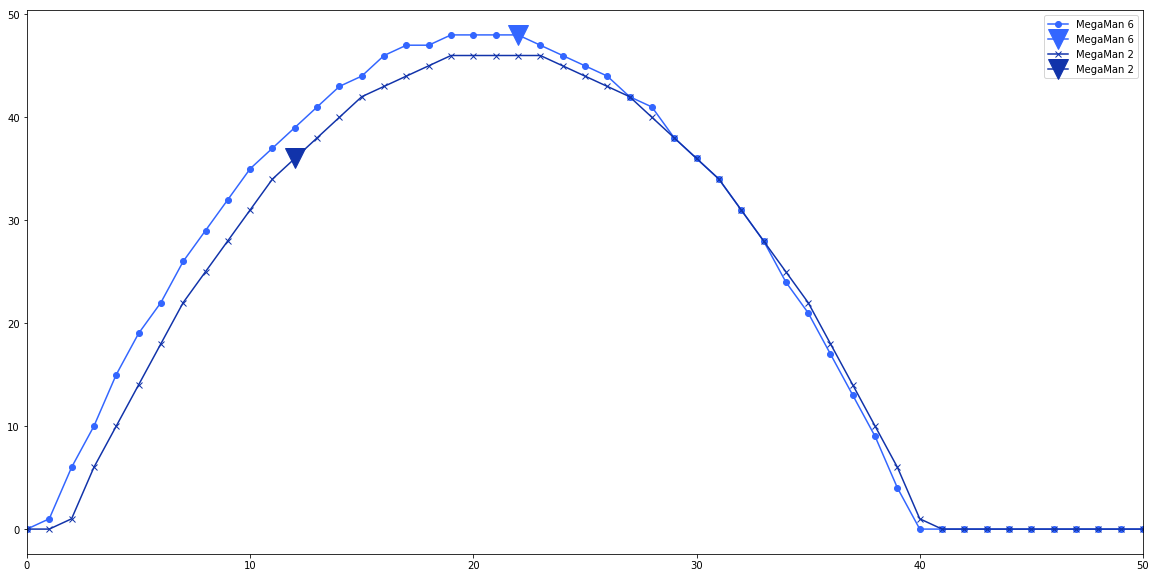}
        \caption{A detailed look at the differences between \emph{Mega Man} 2 and 6.  The triangles represent the \textit{Max Hold Duration}, i.e. if they player had released the button before that point the jumps would be different.  }
        \label{fig:MegaMan}
\end{figure}

\begin{figure}[h!]
        \centering
        \includegraphics[width=0.46\textwidth]{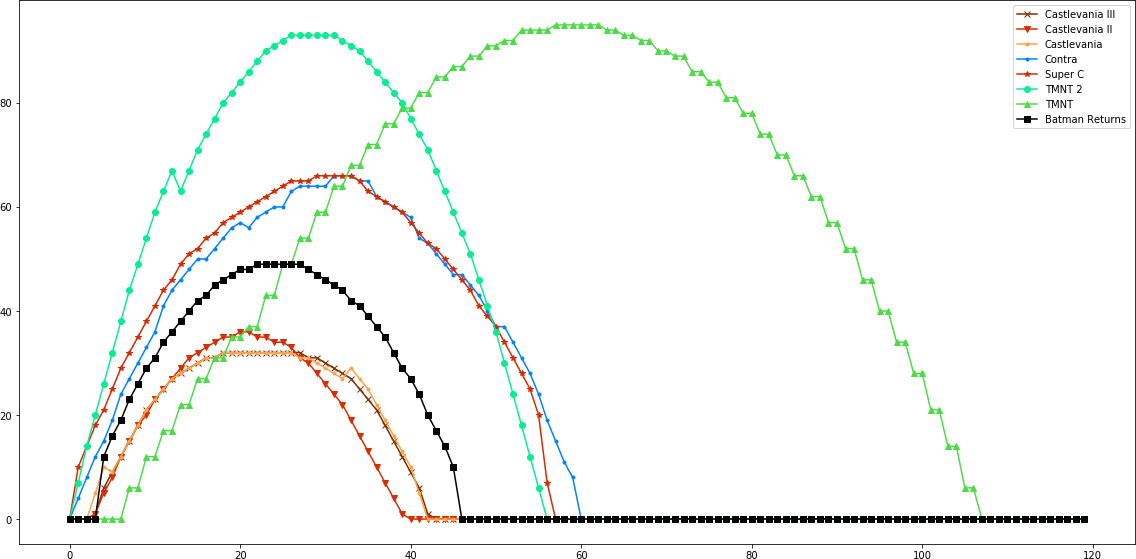}
        \caption{A comparison of jumps across games developed by Konami on the NES.  }
        \label{fig:KonamiComp}
\end{figure}
While the quantitative analyses are valuable, we can also perform a qualitative analysis to compare jumps across different games within a series or by a developer.  Figure \ref{fig:MarioComp} shows all of the jumps in a Mario game on the NES.  We note that no jumps in the series are exactly the same, perhaps most surprisingly with \emph{Mario Is Missing!} and \emph{Mario's Time Machine}, two late era games that used modified \emph{Super Mario World} graphics focused on edutainment.  Two pairs stand out as being more similar than the rest, (1) Toad and Mario in \emph{Super Mario Bros. 2} and (2) \emph{Super Mario Bros.} and \emph{Super Mario Bros. 3}.  The first is surprising since the manual states,

\begin{displayquote}
\hspace{-0.5cm}\textbullet Mario

\hspace{-0.5cm}Average Jumping power in almost all situations.

\noindent \hspace{-0.5cm}\textbullet Toad

\hspace{-0.5cm}He has the least jumping power.  \cite{smb2manual}
\end{displayquote}
 
\noindent when their jumps differ by at most 2 pixels.  \emph{Super Mario Bros.} and \emph{Super Mario Bros. 3} being very similar is somewhat surprising, given the length of time between the games and the large differences between both and \emph{Super Mario Bros. 2}. This can potentially be explained by the original release of Super Mario Bros. 2 in Japan, which retained almost identical mechanics to Super Mario Bros., with additions for added challenge. However, we were unable to do an analysis for that title, as test runs resulted in abnormal data; this may be a feasible point to explore in further work, as it would resolve the discrepancies among these titles.

Turning our attention to games developed by Nintendo (figure \ref{fig:NESComp}) we see no traits that would indicate a cohesive jumping design philosophy.  \emph{Metroid}, \emph{Kid Icarus} and  \emph{Kirby} all have standard parabolas, indicating that the gravity while rising is the same as the gravity while falling.  This is interesting, since \emph{Super Mario Bros.}'s stronger gravity while falling is one of its defining characteristics.  Perhaps unsurprising given its more realistic proportions, \emph{Zelda 2} has one of the smallest jumps, second only to \emph{Balloon Fight}, which has a very small jump with very low gravity.

Looking at games developed by Capcom, we see that despite all \emph{Mega Man} games sharing the same graphics, they all differ in their jumps by a few pixels, but are timed so that they all land within a frame of each other. This can potentially be explained by a change in the development team between Mega Man 2 and Mega Man 3; Akira Kitamura directed the first two games before leaving the company, and Capcom assigned Masahiko Kurokawa, a programmer who had worked on \emph{Chip 'n Dale's Rescue Rangers} \cite{mm3bossfight}. 
While it would have been possible to reuse jump code for the licensed games, both \emph{Chip 'n Dale's Rescue Rangers} and \emph{Duck Tales} have unique jumps.  \emph{Darkwing Duck} has a unique jump, but shares many of the same features as the \emph{Mega Man} games, \textit{hold duration}, \textit{jump duration}, and very similar reset and gravity values. This situation affirms our initial hypothesis that one can make use of feature analysis to ensure consistency of game feel. We also note that \emph{Ghosts 'n Goblins} follows a very non-physical jump with a stair-case pattern where Arthur spends 2-4 frames at a height and then jerkily snaps to the next height.  Our model performs poorly on jumps like this, due to our assumption that the jumps will be governed by in game physics.

While the maximum jump arcs for the \textit{Mega Man} series all appear very similar, differing in height by a few pixels and all landing within a frame of each other, the jump parameters tell a different story. Figure \ref{fig:MegaMan} shows a comparison of the two different types of jumps found in the series.  The first two games have a \textit{max hold} of 12 frames, while the latter games have a \textit{max hold} nearly double that of 20 frames.  By going from a fifth of a second up to a third of a second, players have finer control over their jumps.  With the lengthening of the jump control period, a change had to be made to the jump dynamics.  As mentioned above, all jumps are roughly equivalent in maximum height and duration, but after the button is released, the first two games reset to a small upwards velocity; However, if the latter games kept this small upwards velocity, they would have had a much higher, longer jump.  Since the max hold for the later games comes at the apex of the longest jump, the release of the button resets the velocity to 0, making the longest jump a perfect parabola.

Finally, figure \ref{fig:KonamiComp} shows a comparison of the games developed by Konami.  While the jumps are not identical, both games in the \emph{Contra} series (\emph{Contra} and \emph{Super C}) share very similar jumps.  \emph{Castlevania} and \emph{Castlevania III} also have very similar jumps, but surprisingly \emph{Castlevania II} has a completely unique jump.  Beyond those two series, there are no real similarities between jumps, especially between \emph{Teenage Mutant Ninja Turtles} and \emph{Teenage Mutant Ninja Turtles II: The Arcade Game} which represents the largest difference in jumps between two games ostensibly in the same series.

Our findings seem to corroborate the potential for creating new entries in franchises with similar, if not identical, game feel by analyzing the parameters and features of jumps and other mechanics of previous titles. Further research could be conducted on the potential for game team "mechanics bibles" similar to design bibles used by artists to maintain art styles across franchises.

\section{Conclusion}

In this paper we have presented a framework that uses a modified emulator to automatically run a series of experiments to determine the form and parameterization of a given game's jump model.  We then applied that framework to a corpus of games over 10 times larger than has been analyzed before, allowing us to perform both quantitative and qualitative analyses to find commonalities and trends across many platform games on the NES.  

There still exists work to be done with this framework, most simply by fully cataloging the platformers on the NES.  Toward that end, we would also like a more general framework for learning when a jump has begun (some games have delays between button press and jump, either fixed or variable in length), when a jump has ended (given the odd landing animations of some games), better handling of stair step jumps, automatic jump button determination, and better handling of animation jitter.  Furthermore, we would like to also learn different aspects of jumps, such as double (or triple or infinite) jumping, as well as different preconditions for jumping (e.g. Mario standing still and running will produce different jumps).  To do this, we would also like to get into these different states in an unguided manner, perhaps using Monte Carlo Tree Search to find safe, experimentally valid states. While jumping is a critical component of platformers, it does not exist in a vacuum.  We would also like to extend this work to determine how jumping interacts with other elements, e.g. Mario bumping his head on a brick or landing on a Goomba. Eventually, we would like to learn mechanics, whole cloth, with minimal human interaction.

Finally, we would like to link the learned mechanical properties back to a key piece of inspiration for this work: \textit{Game Feel}.  Swink has some loose rules about the game feel associated with certain parameterizations (low gravity feels floaty, lack of jump control feels limiting), but we hope that this work can enable a larger methodical study to move beyond intuition.


\bibliographystyle{ACM-Reference-Format}
\bibliography{bibliography} 

\end{document}